\documentclass[11pt]{article}
\usepackage[a4paper]{geometry}
\usepackage{clic2017}
\usepackage{times}
\usepackage{url}
\usepackage{latexsym}
\usepackage{float}
\usepackage{flexisym}
\def \name {BERTino}

\title{\name: an Italian DistilBERT model}

\author{Matteo Muffo \\
  Indigo.ai \\ Via Torino 61, Milano \\
  {\tt matteo@indigo.ai} \\\And
  Enrico Bertino \\
  Indigo.ai \\ Via Torino 61, Milano \\
  {\tt e@indigo.ai} \\}

\date{}

\begin{document}
\maketitle
\begin{abstract}
\textbf{English.}\footnote{Copyright \textcopyright2020 for this paper by its authors. Use permitted under Creative Commons License Attribution 4.0 International (CC BY 4.0).} The recent introduction of Transformers language representation models allowed great improvements in many natural language processing (NLP) tasks. However, if on one hand the performances achieved by this kind of architectures are surprising, on the other their usability is limited by the high number of parameters which constitute their network, resulting in high computational and memory demands. In this work we present \name, a DistilBERT model which proposes to be the first lightweight alternative to the BERT architecture specific for the Italian language. We evaluated \name\ on the Italian ISDT, Italian ParTUT, Italian WikiNER and multiclass classification tasks, obtaining F1 scores comparable to those obtained by a $BERT_{BASE}$ with a remarkable improvement in training and inference speed.

\textbf{Italiano.} La recente introduzione dei Transformers come modelli di rappresentazione del linguaggio naturale ha permesso grandi avanzamenti sullo stato dell'arte in molte applicazioni di Natural Language Processing (NLP). Tuttavia, se da una parte i risultati raggiunti da queste architetture sono sorprendenti, dall'altra la loro fruibilità è limitata dall'elevato numero di parametri che costituiscono la loro architettura, con conseguenti elevate esigenze computazionali e di memoria. In questo lavoro presentiamo \name, un modello DistilBERT che è la prima alternativa \textit{leggera} all'architettura BERT specifica per la lingua italiana. Abbiamo valutato \name\ sui task ISDT italiano, ParTUT italiano, WikiNER italiano e classificazione multiclasse, ottenendo punteggi F1 paragonabili a quelli ottenuti da un modello $ BERT_ {BASE} $ con un notevole miglioramento nella velocità di addestramento e inferenza.
\end{abstract}

\section{Introduction} 
\label{sec:introduction}

In recent years the introduction of Transformers language models allowed great improvements in many natural language processing (NLP) tasks. Among Transformer language models, BERT \cite{devlin2018bert} affirmed itself as an high-performing and flexible alternative, being able to transfer knowledge from general tasks to downstream ones thanks to the pretraining-finetuning approach. The context-dependent text representations provided by this model demonstrated to be a richer source of information when compared to static textual embeddings such as Word2Vec \cite{mikolov2013efficient}, GloVe \cite{pennington-etal-2014-glove}, FastText \cite{bojanowski2016enriching} or Sent2Vec \cite{Pagliardini_2018}. However, despite the substantial improvements brought by BERT in the NLP field, the high number of parameters that constitute its network makes its usage prohibitive in resource-limited devices, both at training and inference time, and with a non-negligible environmental impact. To address the aforementioned problem, recent research proposes several approaches to reduce the size of the BERT network, such as DistilBERT \cite{sanh2019distilbert}, MobileBERT \cite{sun2020mobilebert} or pruning \cite{gordon2020compressing,mccarley2019structured}.

The experiments conducted in \newcite{finnish}, \newcite{vries2019bertje} and \newcite{camembert} demonstrate that monolingual BERT models outperform the same multilingual BERT architecture \cite{devlin2018bert}, justifying the effort for pre-training Transformer models required for specific languages.
In this work we present \textbf{\name}, a DistilBERT model pre-trained on a large Italian corpus. This model proposes to be the first general-domain, lightweight alternative to BERT specific for the Italian language. We evaluate \name\ on two Part Of Speech tagging tasks, Italian ISDT \cite{bosco-isdt} and Italian ParTUT \cite{partut}, on the Italian WikiNER \cite{wikiner} Named Entity Recognition task and on a multi-class sentence classification. Comparing the scores obtained by BERTino, its teacher model and GilBERTo, the first obtains performances comparable to the other two architectures while sensibly decreasing the fine-tuning and evaluation time. In Section 2 we discuss the related works with a focus on DistilBERT, in Section 3 we describe the corpus and the pre-train followed by the results in Section 4.

\section{Related work} \label{sec:related_work}
In this section we will give a brief outline of the inner workings for Transformers, then we overview some lightweight alternatives to BERT.

The introduction of Transformer blocks \cite{vaswani2017attention} in language representation models is a keystone in recent NLP. The attention mechanism adopted by the Transformer encoder allows to provide contextualized representations of words, which proved to be a richer source of information than static word embeddings. Attention mechanism processes all words in an input sentence simultaneously, allowing parallelization of computations. This is a non-negligible improvement with respect to models like ELMo \cite{peters2018deep}, which aim to provide contextualized text representations using a bidirectional LSTM network, processesing each word sequentially.

Among language models that adopt Transformer technology, BERT \cite{devlin2018bert} affirmed itself as a flexible and powerful alternative, being able to establish  new state-of-the-art for 11 NLP tasks at the time of publication. In its base version, this model adopts an hidden size of 768 and is composed of 12 layers (Transformer blocks), each of these involving 12 attention heads, for a total of 110 millions of parameters. As outlined in Section \ref{sec:introduction}, the high number of parameters constituting BERT's network can result prohibitive for deployment in resource-limited devices and the computational effort is not negligible. For this reason, great effort has been devoted by researchers in order to propose smaller but valid alternatives to the base version of BERT. \newcite{gordon2020compressing} studies how weight pruning affects the performances of BERT, concluding that a low level of pruning (30-40\% of weights) marginally affects the natural language understanding capabilities of the network.

\newcite{mccarley2019structured} conducts a similar study on BERT weight pruning, but applied to the Question Answering downstream task specifically.

\newcite{sanh2019distilbert} propose DistilBERT, a smaller BERT architecture which is trained using the knowledge distillation technique \cite{hinton2015distilling}. Since the model that we propose relies on this training technique, we propose a brief description of knowledge distillation in section \ref{sec:kd}. DistilBERT leverages the inductive biases learned by larger models during pre-training using a triple loss combining language modeling, distillation and cosine-distance losses. DistilBERT architecture counts 40\% less parameters but is able to retain 97\% of natural language understanding performances with respect to the teacher model, while being 60\% faster. 

\newcite{sun2020mobilebert} propose MobileBERT, a compressed BERT model which aims to reduce the hidden size instead of the depth of the network. As DistilBERT, MobileBERT uses knowledge distillation during pre-training but adopts a $BERT_{LARGE}$ model with inverted bottleneck as teacher.

\subsection{Knowledge distillation} \label{sec:kd}
Knowledge distillation \cite{hinton2015distilling} is a training technique that leverages the outputs of a big network (called \textit{teacher}) to train a smaller network (the \textit{student}). In general, in the context of supervised learning, a classifier is trained in such a way that the output probability distribution that it provides is as similar as possible to the one-hot vector representing the gold label, by minimizing the cross-entropy loss between the two. By receiving a one-hot vector as learning signal, a model evaluated on the training set will provide an output distribution with a near-one value in correspondence of the right class, and all near-zero values for other classes. Some of the near-zero probabilities, however, are larger than the others and are the result of the generalization capabilities of the model. The idea of knowledge distillation is to substitute the usual one-hot vector representing gold labels with the output distribution of the teacher model in the computation of the cross-entropy loss, in order to leverage the information contained in the near-zero values of the teacher's output distribution. Formally, the knowledge distillation loss is computed as:
\begin{equation} \label{eq:kd_loss}
    \mathcal{L}_{KD} = \sum_i t_i * \log(s_i)
\end{equation}
with $t_i$ being the output distribution of the teacher model relative to the $i^{th}$ observation, and $s_i$ being the output distribution of the student model relative to the $i^{th}$ observation.

\section{\name} \label{sec:model}
As outlined in section \ref{sec:introduction}, we propose in this work \name, a DistilBERT model pre-trained on a general-domain Italian corpus. As for BERT-like architectures, \name\ is task-agnostic and can be fine-tuned for every downstream task. In this section we will report details relative to the pre-training that we conducted.

\subsection{Corpus}

The corpus that we used to pre-train \name\ is the union of PAISA \cite{lyding-etal-2014-paisa} and ItWaC \cite{itwac}, two general-domain Italian corpora scraped from the web. While the former is made up of short sentences, the latter includes a considerable amount of long sentences. Since our model can receive input sequences of at most 512 tokens, as for BERT architectures, we decided to apply a pre-processing scheme to the ItWaC corpus. We split the sentences with more than 400 words into sub-sentences, using fixed points to create chunks that keep the semantic sense of a sentence. In this way, most of the long sentences contained in ItWaC are split into sub-sentences containing less than 512 tokens. A certain number of the final sentences still contain more than 512 tokens and they will be useful for training the parameters relative to the last entries of the network.

The PAISA corpus counts 7.5 million sentences and 223.5 million words. The ItWaC corpus counts 6.5 million sentences and 1.6 billion words after preprocessing. Our final corpus counts 14 million sentences and 1.9 billion words for a total of 12GB of text.

\subsection{Pre-training}

\textbf{Teacher model} The teacher model that we selected to perform knowledge distillation during the pre-training of \name\ is \textit{dbmdz/bert-base-italian-xxl-uncased}, made by \textit{Bavarian State Library}\footnote{https://github.com/dbmdz/berts}. We chose this model because it is the Italian $BERT_{BASE}$ model trained on the biggest corpus (81 GB of text), up to our knowledge. Following \newcite{sanh2019distilbert}, we initialized the weights of our student model by taking one layer out of two from the teacher model. 

\textbf{Loss function} We report the loss function used to pre-train \name:
\begin{equation} \label{eq:loss}
    \mathcal{L}=0.45\mathcal{L}_{KD} + 0.45\mathcal{L}_{MLM} + 0.1\mathcal{L}_{COS}
\end{equation} with $\mathcal{L}_{KD}$ being the knowledge distillation loss as described in equation \ref{eq:kd_loss}, $\mathcal{L}_{MLM}$ being the masked language modeling loss and $\mathcal{L}_{COS}$ being the cosine embedding loss. \newcite{sanh2019distilbert} describe the cosine embedding loss useful to ``align the directions of the student and teacher hidden states vectors''. When choosing the weights of the three loss functions, we wanted our model to learn from the teacher and by itself in an equal way, so we set the same weights for both $\mathcal{L}_{KD}$ and $\mathcal{L}_{MLM}$. Moreover, we considered the alignment of student and teacher hidden states vectors marginal for our objective, setting $\mathcal{L}_{COS}$ as 10\% of the total loss.

\textbf{Architecture} The architecture of \name\ is the same as in DistilBERT. Our model adopts an hidden size of 768 and is composed of 6 layers (Transformer blocks), each of which involving 12 attention heads. In this way \name's network results to have half the layers present in the $BERT_{BASE}$ architecture.

\textbf{Training details} To pre-train \name\ we used a batch size of 6 and an initial learning rate of $5 \times 10^{-4}$, adopting Adam \cite{kingma2014adam} as optimization algorithm. We chose 6 as batch size due to the limited computational resources available. Results described in section \ref{sec:results} demonstrate that the small batch size that we adopted is sufficient to obtain a valid pre-trained model. We trained our model on 4 Tesla K80 GPUs for 3 epochs, requiring 45 days of computation in total. For some aspects of the training, we relied on the Huggingface Transformers repository \cite{Wolf2019HuggingFacesTS}.

\section{Results} \label{sec:results}

We tested the performances of \name\ on benchmark datasets: the Italian ISDT \cite{bosco-isdt} and Italian ParTUT \cite{partut} Part Of Speech tagging tasks, and the Italian WikiNER \cite{wikiner} Named Entity Recognition task. To complete the evaluation of the model, we also tested it on a multi-class sentence classification task. In particular, we focused on intent detection, a task specific to the context of Dialogue Systems, creating a novel italian dataset which is freely available at our repository\footnote{https://github.com/indigo-ai/BERTino}. The dataset that we propose collects 2786 real-world questions (2228 for training and 558 for testing) submitted to a digital conversational agent. The total number of classes in the dataset is 139.

For the first two tasks mentioned, we fine-tuned our model on the training set for 4 epochs with a batch size of 32 and a learning rate of $5 \times 10^{-5}$, for the NER task we performed 5-fold splitting of the dataset and fine-tuned \name\ for 2 epochs per fold with a batch size of 32 and a learning rate of $5 \times 10^{-5}$, while for the multi-class classification task  we fine-tuned our model for 14 epochs on the training set with a batch size of 32 and a learning rate of $5 \times 10^{-5}$. To compare the results obtained, we fine-tuned the teacher model and a GilBERTo model\footnote{Available at https://github.com/idb-ita/GilBERTo} on the same tasks with the same hyperparameters. Tables \ref{tab:isdt}, \ref{tab:partut}, \ref{tab:ner} and \ref{tab:multiclass} collect the F1 scores gathered in these experiments together with fine-tuning and evaluation time. All the scores reported represent the average computed over three different runs. Results show that the teacher model slightly outperforms \name, with an increase of the F1 score of 0,29\%, 5,15\%, 1,37\% and 1,88\% over the tasks analysed. However \name\ results to be a sensibly faster network with respect to the teacher model and GilBERTo, taking almost half of the time to perform both fine-tuning and evaluation. We can conclude from the last observation that \name\ is able to retain most of the natural language understanding capabilities of the teacher model, even with a much smaller architecture.

\begin{table*}[bpt]
    \centering
    \begin{tabular}{|l|c|c|c|}
         \hline
         \multicolumn{4}{|c|}{Italian ISDT} \\
         \hline
         Model & F1 score & Fine-tuning time & Evaluation time \\
         \hline
         \name & 0,9800 & 9'10'' & 3''  \\
         Teacher model & 0,9829 & 16'32'' & 6'' \\
         GilBERTo & 0,9804 & 18'11'' & 5'' \\
         \hline
    \end{tabular}
    \caption{F1 scores obtained by \name\ and the teacher model in the Italian ISDT task.}
    \label{tab:isdt}
\end{table*}

\begin{table*}[bpt]
    \centering
    \begin{tabular}{|l|c|c|c|}
         \hline
         \multicolumn{4}{|c|}{Italian ParTUT} \\
         \hline
         Model & F1 score & Fine-tuning time & Evaluation time \\
         \hline
         \name & 0,9193 & 1'19'' & 1''  \\
         Teacher model & 0,9708 & 2'19'' & 1''  \\
         GilBERTo & 0,9621 & 2'21'' & 1''  \\
         \hline
    \end{tabular}
    \caption{F1 scores obtained by \name\ and the teacher model in the Italian ParTUT task.}
    \label{tab:partut}
\end{table*}

\begin{table*}[hbpt]
    \centering
    \begin{tabular}{|l|c|c|c|}
         \hline
         \multicolumn{4}{|c|}{Italian WikiNER} \\
         \hline
         Model & F1 score & Fine-tuning time & Evaluation time \\
         \hline
         \name & 0,9039 & 38'3'' & 3'2''   \\
         Teacher model & 0,9176 & 67'2'' & 5'21''  \\
         GilBERTo & 0,9136  & 66'33''  & 5'9''  \\
         \hline
    \end{tabular}
    \caption{F1 scores obtained by \name\ and the teacher model in the Italian WikiNER task. The results reported are the average of the scores obtained in each of the 5 folds.}
    \label{tab:ner}
\end{table*}

\begin{table*}[hbpt]
    \centering
    \begin{tabular}{|l|c|c|c|}
         \hline
         \multicolumn{4}{|c|}{Multi-class sentence classification} \\
         \hline
         Model & F1 score & Fine-tuning time & Evaluation time \\
         \hline
         \name & 0,7766 & 5'4'' & 6''  \\
         Teacher model & 0,7954 & 9'48'' & 10''  \\
         GilBERTo & 0,7381  & 10'0'' & 10''  \\
         \hline
    \end{tabular}
    \caption{F1 scores obtained by \name\ and the teacher model in the multi-class sentence classification task.}
    \label{tab:multiclass}
\end{table*}

\section{Conclusions}

In this work we presented \name, a DistilBERT model which aims to be the first lightweight alternative to BERT specific for the Italian language. Our model has been trained on a general-domain corpus and can then be finetuned with good performances on a wide range of tasks like its larger counterparts. \name\ showed comparable performances with respect to both the teacher model and GilBERTo in the Italian ISDT, Italian ParTUT, Italian WikiNER and multi-class sentence classification tasks while taking almost half of the time to fine-tune, demonstrating to be a valid lightweight alternative to $BERT_{BASE}$ models for the Italian language.
\newpage
\bibliographystyle{acl}
\bibliography{bibliography}

\end{document}